\documentclass{article}

\usepackage{arxiv}

\usepackage[utf8]{inputenc} 
\usepackage[T1]{fontenc}    
\usepackage{hyperref}       
\usepackage{url}            
\usepackage{booktabs}       
\usepackage{amsfonts}       
\usepackage{nicefrac}       
\usepackage{microtype}      
\usepackage{lipsum}		
\usepackage{graphicx}
\usepackage{natbib}
\usepackage{doi}

\usepackage{xcolor}
\usepackage{amssymb}
\usepackage{bm}
\usepackage{subfigure}
\usepackage{amsmath}
\usepackage[most]{tcolorbox}

\title{Demystifying Design Choices of Reinforcement Fine-tuning: 
A Batched Contextual Bandit Learning Perspective}

\author{
Hong Xie\textsuperscript{1}\quad
Xiao Hu\textsuperscript{1}\quad
Tao Tan\textsuperscript{1}\thanks{Corresponding author. Email: \texttt{tantao1222@ustc.edu.cn}}\quad
Haoran Gu\textsuperscript{2}\quad
Xin Li\textsuperscript{3}\quad
Jianyu Han\textsuperscript{3}\quad
Defu Lian\textsuperscript{1}\quad
Enhong Chen\textsuperscript{1}\quad
\\[4pt]
\textsuperscript{1}University of Science and Technology of China\quad
\textsuperscript{2}Daqing Oilfield Chongqing Company\quad
\textsuperscript{3}IFlyTek (China)
}

\date{}

\hypersetup{
pdftitle={Demystifying Design Choices of Reinforcement Fine-tuning: 
A Batched Contextual Bandit Learning Perspective},
pdfauthor={Xiao Hu, Hong Xie, Tao Tan, Defu Lian, Jianyu Han},
}

\begin{document}
\maketitle

\begin{abstract}
The reinforcement fine-tuning area is undergoing an explosion papers 
largely on optimizing design choices.  
Though performance gains are often claimed, 
inconsistent conclusions also arise from time to time, 
making the progress illusive.   
Reflecting on this illusion, 
we still lack principled answers to two fundamental questions: 
1) what is the role of each design choice? 
2) which ones are critical?  
This paper aims to shed light on them.   
The underlying challenge is that design choices are entangled together,  
making their contribution to learning and generalization difficult to attribute.   
To address this challenge, we first construct a minimalist baseline 
for disentangling factors: 
one rollout per query in each round, 
the outcome reward  serving as the training signal 
without any advantage trick, and a batch size of thirty-two.   
This baseline connects to 
batched contextual bandit learning, 
which facilitates experimental analysis.  
Centering around this baseline, 
we design an experiment pipeline, 
examining the marginal gains of factors like advantage, number of rollouts, etc.  
Experiments on three base models and two datasets,  
not only reveal new understanding on the role of various design choices 
on learning and generalization dynamics, 
but also identify critical ones that deserve more effort. 
\end{abstract}

\section{Introduction}
\label{sec:intro}
The reinforcement fine-tuning area is undergoing an explosion of papers 
largely focused on optimizing design choices from the perspective of reinforcement learning (RL) \cite{Wang2025,Zhang2025a}.   
Tracing the vast literature back, 
it is rooted in a policy view of LLMs, 
which enabled the application of policy-based algorithms like 
PPO \cite{schulman2017proximal,ouyang2022training,ChatGPT22} to fine-tune them.  
Most of the literature concentrates on critic-free policy algorithms, 
because they are more computationally efficient, 
where DPO \cite{Rafailov2023} and GRPO \cite{shao2024deepseekmath} 
are two representative ones.  
Drawing intuitions from RL, 
various design choices of reinforcement fine-tuning were optimized, 
such as credit assignment mechanism \cite{Kazemnejad2025}, 
entropy regularization \cite{Wang2025}, advantage function \cite{li2024remax}, 
gradient optimization \cite{yu2025dapo}, etc.

Though performance gains are often claimed, 
inconsistent conclusions also arise from time to time, 
making the progress elusive.    
First, we take entropy regularization as an example \cite{Wang2025}.  
Several studies designed regularizers to encode the intuition 
from classical RL that higher entropy promotes exploration, 
and claimed generalization improvement 
\cite{Entropy-Perspective25,GTPO&GRPO-S25}.   
The opposite view that lower entropy leads to better generalization 
was also claimed \cite{Entropy-Minimization25}.  
Furthermore, many works optimized the advantage function and claimed 
better generalization \cite{Zhang2025a}.  
But recently, random rewards are even shown to improve 
the generalization of base models like Qwen \cite{Shao2025}.

Reflecting on the elusive progress, 
two reasons account for it.  
First, though the policy view of LLMs connects  
reinforcement fine-tuning with RL, 
the essential difference is far from clear.  
Most efforts were spent on showcasing the positive results  
of applying ideas from RL, 
instead of figuring out when they work and when they do not.   
Second, principled answers to two fundamental questions are lacking: 
1) what is the role of each design choice? 
2) which ones are critical?  
Several works have attacked these problems from an empirical perspective \cite{Jin2025,Swamy2025,Tan2025} 
and a theoretical perspective \cite{Chen2025,Dylan2025,Xie2025,Zhao2025}. 
Though a number of novel insights have been yielded, 
we are still far from satisfactory answers.  
Empirical studies have limitations in generalizing observations 
across different base models and datasets.  
Theoretical studies fall short in practice, 
because they discard important factors of LLMs to make the model 
theoretically tractable.

This paper aims to shed light on the aforementioned questions.   
The underlying challenge is that design choices are entangled together,  
making their contribution to learning and generalization difficult to attribute \cite{Wang2025,Zhang2025a}.   
To address it, we start from baseline construction.  
Inspired by the first principles, the baseline should contain as few factors as possible, 
but capture the nature of reinforcement fine-tuning.  
To the best of our efforts, we construct a minimalist baseline:  
one rollout per query in each round, 
the outcome reward serving as the learning signal 
without any advantage trick, and a batch size of thirty-two.   
This baseline connects to 
batched contextual bandit learning \cite{Jiang2025,Ruan2021}, 
which offers theoretical insights such as batch-regret tradeoffs 
for understanding the experimental results.    
Centering around this baseline, 
we design an experimental pipeline, 
examining the marginal gains of factors like advantage, number of rollouts, etc.  
Experiments on three base models and two datasets,  
not only reveal new insights into the role of various design choices in learning and generalization dynamics, 
but also identify critical ones that deserve more effort. 
Note that the implementation of this work can be found at 
\url{https://github.com/tt725/Bandit-Reinforcement-Fine-tuning.git}.

\subsection{Contributions}

\noindent
{\bf Experiment pipeline.}  
An experiment pipeline inspired by batched contextual bandits, 
which examines the marginal gains of factors like advantage, number of rollouts, etc.  

\noindent
{\bf Critical experimental findings.}
Several new experimental findings are observed across three base models and two dataset pairs:  
1) The Pass@1 over training and test data increases at comparable rates as the number of 
training rounds increases;  
2) Under the minimalist baseline, the improvement of Pass@1 over the test data varies from $0.00$ to $0.60$;  
3) By enabling GRPO-type advantage, the marginal improvement of Pass@1 over the test data varies from $0.00$ to $0.10$;  
4) By further increasing the number of rollouts from $1$ to $64$, the marginal improvement of Pass@1 over the test data varies from $0.04$ to $0.17$; 
5) By scaling the batch size from $32$ to $128$ when the number of rollouts equals eight, the marginal improvement of Pass@1 over the test data varies from $0.00$ to $0.05$; 
6) For a fixed budget, there is a tradeoff between batch size and number of rollouts, and a replay strategy is designed to attain the optimal tradeoff; 
7) Scaling batch size has diminishing returns in the replay strategy, and its ceiling is revealed.

\noindent
{\bf Insights for shaping the area.} 
For each heuristic (i.e., optimizing advantage, number of rollouts, batch size, etc.), 
there are model-dataset pairs that lead to nearly no marginal improvement of Pass@1 over the minimalist baseline.   
The area should pay more attention to understanding why some model-dataset pairs generalize well and some do not.  
Furthermore, more efforts should be devoted to building more systematic and unified benchmarks.

\section{Related Work}

\noindent 
{\bf Optimizing advantage function.}  
Which RL algorithm should be used for fine-tuning LLMs has long been debated.  
Three research threads dominate:
1) Actor-critic framework, i.e., PPO \cite{schulman2017proximal,ouyang2022training} and VAPO \cite{yue2025vapo}.
They supply fine-grained value estimates, but the extra value-network inflates GPU memory.
2) Pure policy-gradient \cite{hsu2020revisiting,grudzien2022mirror}, i.e., ReMax \cite{li2024remax}, REINFORCE \cite{ahmadian2024back}, REINFORCE++ \cite{hu2025reinforce++}.  
Eliminating the value-network relieves GPU memory pressure, 
yet the high variance of policy gradient makes training unstable.
3) Advantage function, i.e., GRPO \cite{shao2024deepseekmath}, DAPO \cite{yu2025dapo}, GSPO \cite{zheng2025group}, Dr.GRPO \cite{liu2025understanding} and TreePO \cite{li2025treepo}.  
They estimate an advantage from multiple rollouts to reduce policy gradient variance.
This paper adopts a bandit view to redesign the advantage function as the reward with only one rollout,
and finds that, 
compared with the reward, the advantage of GRPO barely helps reinforcement fine-tuning.
This implies that the advantage function is not a central concern.

\noindent 
{\bf Scaling the number of rollouts.} 
Rollouts, i.e., the number of responses generated per prompt in each step,
have been studied along three main lines in reinforcement fine-tuning:
1) More rollouts yield a better reward estimate and lower policy gradient variance, stabilizing reinforcement fine-tuning \cite{zeng2025shrinking}.
2) Rollout budgets should be allocated according to prompt difficulty to maximize sample efficiency \cite{yang2025depth,qu2025can,sun2025improving}.
3) What matters is not sheer quantity but diversity: varied rollouts keep policy entropy high and prevent collapse \cite{xu2025not,bai2025m,wang2025beyond,cui2025entropy}.
This paper shows that scaling the number of rollouts does not always yield significant performance gains, 
but instead that in some model-dataset pairs the improvement is marginal.   
This implies that the area should shift its focus toward understanding these negative cases.

\noindent 
{\bf Scaling batch size.} 
Batch size, i.e., the number of prompts used to compute policy gradient at each step,
is a key variable in reinforcement fine-tuning,
and it is no longer merely a conventional hyperparameter as in classical RL \cite{parthasarathy2024ultimate,bartoldson2025trajectory,wang2025reinforcement}. 
Current research on the role of batch size in reinforcement fine-tuning can be grouped into three lines:
1) Larger batches stabilize reinforcement fine-tuning \cite{yang2025depth}, which argues that increasing batch size reduces policy gradient variance and thus improves stability.
2) Batch size is a critical scaling factor for reinforcement fine-tuning, and empirical scaling laws have been proposed \cite{khatri2025art}.
3) There is a tradeoff between batch size and GPU memory consumption \cite{gao2025prompt,wu2025llamarl,fu2025areal,sheng2025laminar,wang2025infinite}.
This paper shows that scaling batch size does not always yield significant performance gains, 
but instead that in some model-dataset pairs the improvement is marginal.  
Furthermore, scaling the batch size has a diminishing-return effect, 
and we also reveal its performance ceiling.

\noindent 
{\bf Utilizing replay buffer.}  
Replay buffer, i.e., replaying the responses that LLMs generated in earlier steps, 
aims to raise rollout utilization and stabilize policy gradient updates. 
Three research threads dominate:
1) Replay as the traditional RL experience buffer, 
which smooths policy gradients and stabilizes reinforcement fine-tuning \cite{wang2025eframe,zhang2025rlep,ma2025stabilizing,bartoldson2025trajectory}.
2) Replay for reinforcement fine-tuning speeding \cite{zhou2025april,sun2025improving,liu2025spec,tang2025towards}.
They mitigate the long-tail data distributions and accelerate reinforcement fine-tuning.
3) Replay for improving reinforcement fine-tuning performance \cite{dou2025improving,zhang2025improving}.
They balance exploration and learning to improve the final reward of reinforcement fine-tuning.
For a fixed budget, this paper reveals 
a tradeoff between batch size and number of rollouts 
and designs a replay strategy to attain the optimal tradeoff.  

\noindent 
{\bf Understanding reinforcement fine-tuning.}  
From an empirical perspective, some works investigated 
the role of RL through the lens of information theory 
\cite{Swamy2025}, the role of data influence \cite{Tan2025}, 
the generalization capability \cite{Jin2025}, etc.  
From a theoretical perspective, models and theories have been established to 
understand the impact of base model \cite{Dylan2025}, 
 limits of outcome-based reward \cite{Chen2025}, 
and sample efficiency \cite{Xie2025,Zhao2025}, etc.  
Though a number of novel insights have been yielded, 
our understanding of reinforcement fine-tuning is still far from clear.  
Empirical studies have limitations in generalizing observations 
across different base models and datasets.  
Theoretical studies fall short in practice, 
because they discard important factors of LLMs to make the model 
theoretically tractable.   
This paper sheds new light on understanding design choices 
of reinforcement fine-tuning.

\section{Background and Problem Statement}

\subsection{Reinforcement Fine-tuning}

We consider reinforcement fine-tuning 
with outcome-level verifiable rewards  $\{0,1\}$,
where $0/1$ represents that a response is incorrect/correct.  
Denote $q$ as a query and denote $\mathcal{O}$ as the response space.  
Let policy $\pi_{\theta}$ denote a base model, 
where $\theta$ denotes the parameters of the base model.  
Policy $\pi_{\theta}$ prescribes 
a probability distribution over $\mathcal{O}$ 
for each query $\pi_{\theta} ( \cdot| q)$.  
More specifically, it holds that 
$\sum_{\bm{y} \in \mathcal{O}} \pi_{\theta}( \cdot| q) = 1$ 
and $\pi_{\theta}( \cdot| q) \in [0,1]$.  
Denote $\mathcal{D}_{train} =\{q_1, \ldots, q_N\}$, 
where $N\in \mathbb{N}_+$, as the dataset for reinforcement fine-tuning.

\subsection{Connections to Batched Contextual Bandits}  

Batched contextual bandits 
are a sequential decision framework to study the 
exploration vs. exploitation tradeoffs under limited 
adaptivity \cite{Jiang2025,Ruan2021,Zhang2021}.  
Each context is associated with a finite (or infinite) number of arms.  
The reward of pulling an arm is jointly determined by the context 
and the pulled arm.  
The reward is usually modeled as a random variable, capturing the uncertainty.  
In each round, a finite number of contexts is revealed to the decision maker in a batch, 
and the decision maker needs to pull one arm for each context.  
Each pulled arm generates a reward, 
which serves as the signal to optimize arm selection policy.   
For each given context, the optimal policy is the one that pulls the arm with the 
largest reward mean.  
The objective is to learn the optimal policy through interactions 
with the environment.  
One typical variant of multi-armed bandits is 
multi-play multi-armed bandits, 
which allows the agent to pull multiple arms.  

The minimalist baseline stated in Section \ref{sec:intro}   
connects to batched contextual bandits as follows.  
Each query can be modeled as a context.  
A batch size of thirty-two can be modeled by the arrival of thirty-two contexts in each decision round.  
One rollout per query corresponds to pulling an arm for each context.  
The verifiable reward can be mapped to the reward generated by an arm.   
The correct response corresponds to the optimal arm.   
Reinforcement fine-tuning simplifies reward function of batched contextual bandits   
in that the reward is deterministic  
and more importantly each reward indicates the optimality of the arm 
(or correctness of the answer).   
It complicates the batched contextual bandits in two aspects: 
(1) the black box base model;  
(2) the focus on generalization performance. 

Batched contextual bandits are simpler than RL in that 
they do not have state transition or delayed consequences.  
In the literature, the reward usually directly serves as the 
signal for policy learning without any advantage design \cite{Jiang2025,Ruan2021,Zhang2021}, 
where fast rates of learning are attained.

\subsection{Problem Statement}
\label{sec:problem}

Formally, the minimalist baseline stated in Section \ref{sec:intro} can be
formally described as follows.  
Let $q_{i,t}$ denote the $i$-th query generated in $t$-th training round, 
where $i \in \{1,\ldots,32\}$.  
Our framework applies to other batch sizes.    
$\bm{o}_{i,t}$ denotes the response for  $q_{i,t}$.    
The outcome reward corresponding to $\bm{o}_{i,t}$ is denoted by $R_{i,t} \in \{0,1\}$.  
The training objective function in round $t$ is a simplification of GRPO: 
\[
\begin{aligned}
\mathcal{J}_t(\theta)
{=}
& 
\!\sum^B_{i=1} \!\!
\sum^{\ell(\bm{o}_{i,t})}_{\tau=1} \!\!
{\min} \Big[\!
r_{i,\tau}(\theta) \!R_{i,t},
\text{clip}\!\left(r_{i,\tau}(\theta), 1{-}\varepsilon, 1{+}\varepsilon\right) \!R_{i,t}
\!\Big]
\\
&
{-} \beta  \mathbb{D}_{\text{KL}}\!\left[\pi_{\theta} \,\|\, \pi_{\text{ref}}\right].
\end{aligned}
\]
Here, $\ell(\bm{o}_{i,t})$ denotes the length of $\bm{o}_{i,t}$, 
and the 
$r_{i,\tau}(\theta) =  \pi_{\theta}(o_{i,t}|q,o_{i,<\tau}) / \pi_{old}(o_{i,\tau}|q,o_{i,<\tau}), \varepsilon$, etc., 
have the same meaning as GRPO \cite{shao2024deepseekmath}.  

Centering around this baseline, 
we design an experimental pipeline, 
examining the marginal gains of factors like advantage, number of rollouts, etc.  
Our objective is to shed light on 
two fundamental questions: 
1) what is the role of each design choice? 
2) which ones are critical?

\section{Experiments}
\label{sec:exp}

\subsection{Experiment Setting}
\label{sec:exp-setting}
We use three representative instruction-tuned LLMs and 
two open-source mathematical reasoning datasets
to study reinforcement fine-tuning.
This results in six model-dataset pairs:
Qwen-GSM, LLaMA-GSM, OLMo-GSM, Qwen-MATH, LLaMA-MATH, and OLMo-MATH.

\textbf{Base Models.}
We use three representative instruction-tuned large language models: Qwen2.5-0.5B-Instruct \cite{Qwen25}, LLaMA-3.2-1B-Instruct \cite{LlaMA3}, and OLMo-2-0425-1B-Instruct \cite{OLMo} 
to study reinforcement fine-tuning. These models represent diverse pretraining and alignment pipelines. Moreover, their relatively small parameter scales make them commonly adopted as lightweight backbones for reinforcement fine-tuning, allowing systematic analysis of learning dynamics and generalization behavior under controlled computational budgets.

\textbf{Datasets.}
Experiments are conducted on two open-source mathematical reasoning benchmarks, MATH \cite{MATH21} and GSM8K \cite{GSM8K21}, using their standard train/test splits for training and evaluation. The MATH dataset contains approximately $7,500$ training problems and $5,000$ test problems spanning multiple difficulty levels and topics. The GSM8K dataset consists of $7,473$ training problems and $1,319$ test problems focused on grade-school mathematical reasoning. These datasets are widely used for evaluating reinforcement fine-tuning methods due to their clear correctness criteria and sensitivity to training stability.

\textbf{Framework.}
We conduct all experiments using \texttt{VeRL}, i.e.,  
\url{https://github.com/verl-project/verl},
fine-tuning framework.
It provides a unified and reproducible infrastructure for reinforcement fine-tuning, supporting GRPO-style policy optimization,
flexible rollout and batching strategies.

\textbf{Learning parameters.}
Unless otherwise specified, we adopt the default \texttt{VeRL} settings, with the following key hyperparameters: 
advantage formulation of GRPO; 
Adam optimizer with a learning rate of $1\times10^{-6}$; 
a KL-divergence coefficient of $0.001$; 
a PPO clipping coefficient of $0.2$; 
maximum question/response lengths of $512/1024$; 
and a micro-batch size of $4$.

\textbf{Metric.}
We use Pass@1 as metric, i.e., 
the fraction of problems for which the model produces a correct solution in its first generated response, 
and record both train Pass@1 and test Pass@1 every $100$ train rounds. For clarity, we define
\begin{align*}
&
\text{train Pass@1 = the Pass@1 evaluated on training dataset}, 
\\ 
& 
\text{test Pass@1 = the Pass@1 evaluated on testing dataset}. 
\end{align*}
During evaluation, responses are generated using the same sampling parameters as those used during training, with temperature 1.0 and top-$p$ 1.0 (i.e., standard softmax sampling). 

\textbf{Hardware and Software.}
All experiments are conducted on a single NVIDIA A100 GPU with 40GB memory.
The software environment is based on Ubuntu~22.04,
with Python~3.12, PyTorch~2.5.1, and CUDA~12.4.
The host machine is equipped with 10 vCPUs
(Intel Xeon Processor, Skylake with IBRS support).

\subsection{Exp1: the Minimalist Baseline}
\label{sec:pipeline}
In this experiment, we study the minimalist baseline stated in Section \ref{sec:problem}.  
It serves as the baseline for later experimental comparison.

\begin{figure}[!htb]
\centering
\begin{minipage}[t]{0.49\linewidth}
\centering
\subfigure{
  \includegraphics[width=0.95\linewidth]{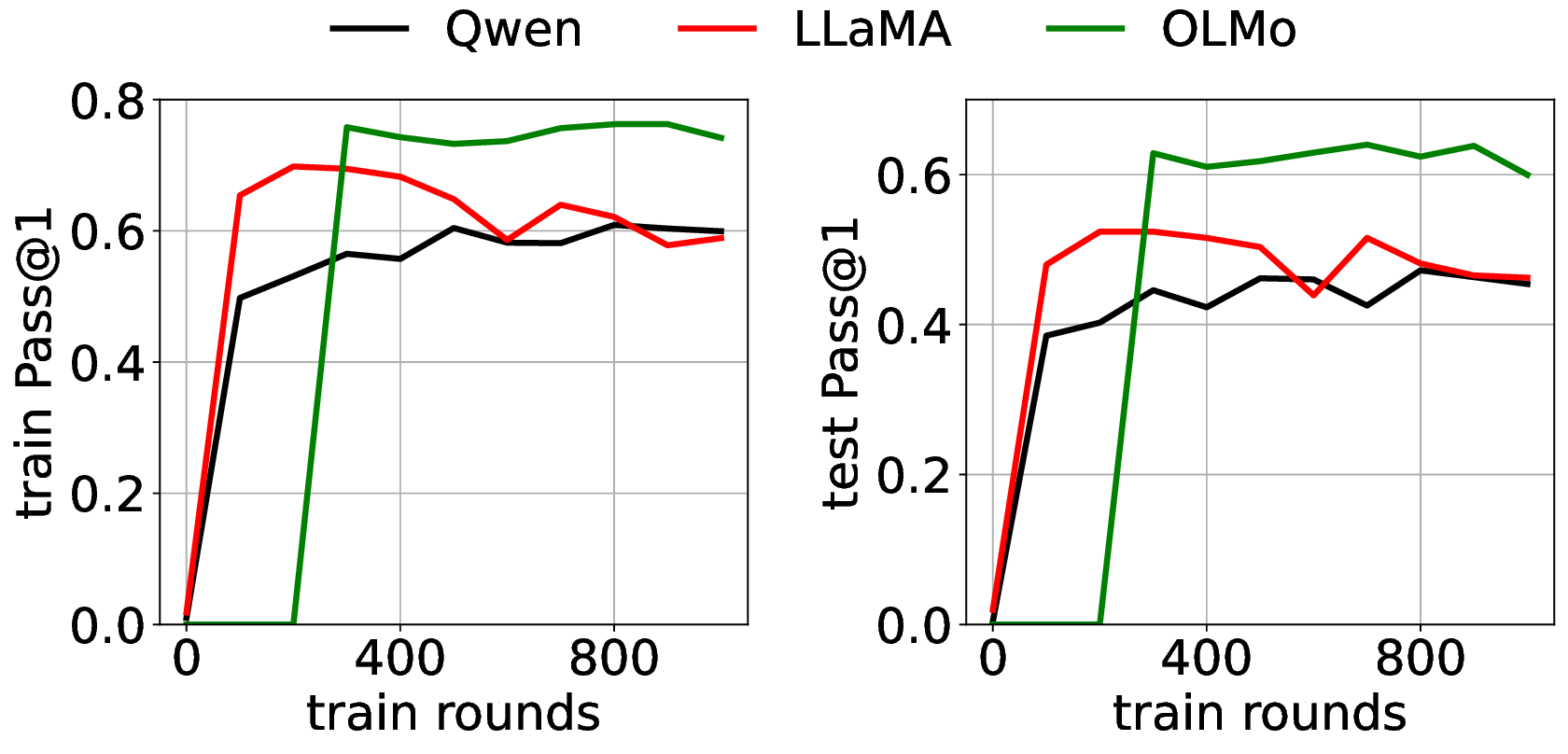}
}
\caption{The minimalist baseline on GSM8K dataset.}
\label{fig:bandit-GSM}
\end{minipage}
\hfill
\begin{minipage}[t]{0.49\linewidth}
\centering
\subfigure{
  \includegraphics[width=0.95\linewidth]{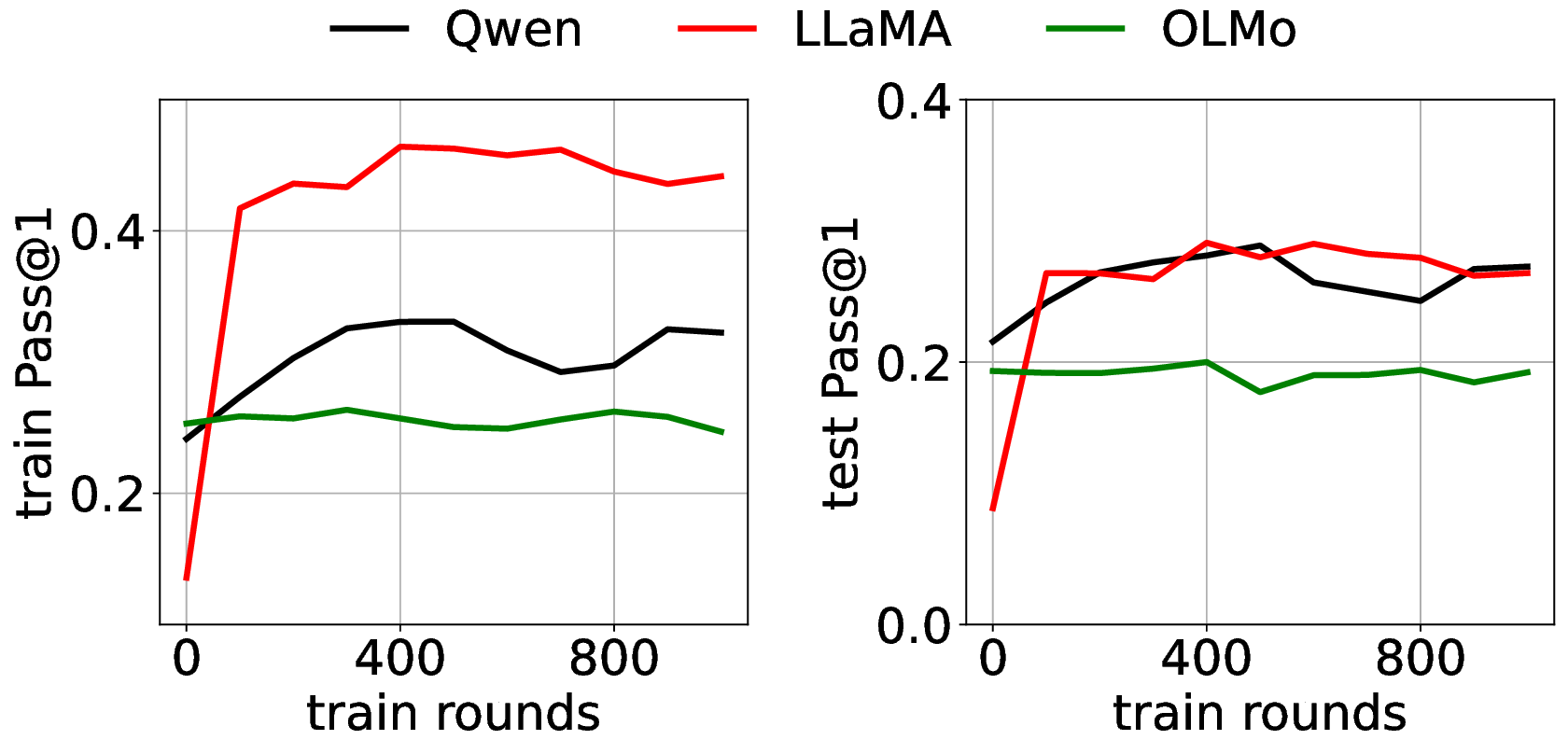}
}
\caption{The minimalist baseline on MATH dataset.}
\label{fig:bandit-MATH}
\end{minipage}
\end{figure}

Figure \ref{fig:bandit-GSM}  
shows the learning and generalization dynamics of the minimalist baseline on GSM8K dataset.
From the learning perspective,
one can observe that,
as the number of training rounds increases, 
the train Pass@1 of three models first increases sharply and then becomes flat.
More specifically, 
the train Pass@1 of Qwen, LLaMA, and OLMo 
improves $0.59$ (from $0.01$ to $0.60$), $0.58$ (from $0.02$ to $0.59$), $0.74$ (from $0.00$ to $0.74$).
Namely, the train Pass@1 is improved by at least $0.57$ across three models.
This implies that under the minimalist baseline,
each base model learns well on GSM8K dataset.
From the generalization perspective,
one can observe that,
as the number of training rounds increases, 
the test Pass@1 of three models also first increases sharply and then becomes flat.
Furthermore,
it increases at the same pace as the train Pass@1,
and at a comparable rate.
More specifically, 
the test Pass@1 of Qwen, LLaMA, and OLMo  
improves by $0.45$ (from $0.00$ to $0.45$), $0.44$ (from $0.02$ to $0.46$), and $0.60$ (from $0.00$ to $0.60$).
Namely, the test Pass@1 is improved by at least $0.44$ across three models.
This implies that under the minimalist baseline,
each base model generalizes well on GSM8K dataset.

Figure \ref{fig:bandit-MATH}  
shows the learning and generalization dynamics of the minimalist baseline on MATH dataset.
From the learning perspective,
it is interesting to observe that
Qwen and LLaMA show clear improvements in train Pass@1, while OLMo shows almost no improvement during training.
More specifically, 
the train Pass@1 of Qwen, LLaMA, and OLMo 
improves by $0.08$ (from $0.24$ to $0.32$), $0.30$ (from $0.14$ to $0.44$), and $\textbf{0.00}$ (from $0.25$ to $0.25$).
From the generalization perspective,
it is interesting to observe that
the test Pass@1 also increases at the same pace as the train Pass@1,
and at a comparable rate.
More specifically, 
the test Pass@1 of Qwen, LLaMA, and OLMo 
improves by $0.05$ (from $0.22$ to $0.27$), $0.18$ (from $0.09$ to $0.27$), and $\textbf{0.00}$ (from $0.19$ to $0.19$).
This implies that the generalization behavior is highly predictable from the learning dynamics.

{\bf Key insights.}  
Under the minimalist baseline:
1) The improvement of test Pass@1 varies from $0.00$ to $0.60$ across six model-dataset pairs,
where OLMo-MATH is $0.00$ (from $0.19$ to $0.19$), and OLMo-GSM is $0.60$ (from $0.00$ to $0.60$).
2) The train Pass@1 of six model-dataset pairs evolves at a comparable pace to the test Pass@1, 
and the generalization behavior is highly predictable from the learning dynamics.

\subsection{Exp2: the Role of Advantage Function}
\label{sec:adv}
We extend the minimalist baseline of Section \ref{sec:pipeline} to study the role of advantage function 
in learning and generalization dynamics.  
To achieve this, we generate $8$ rollouts to obtain $8$ rewards
and compute the advantage function following GRPO \cite{shao2024deepseekmath}.
Note that to ensure a fair comparison,
only the advantage corresponding to the first rollout is used to compute the policy gradient, 
while the remaining rollouts are used to support the advantage of the first rollout.

\begin{figure}[!htb]
\centering
\begin{minipage}[t]{0.49\linewidth}
\centering
\subfigure{
  \includegraphics[width=0.95\linewidth]{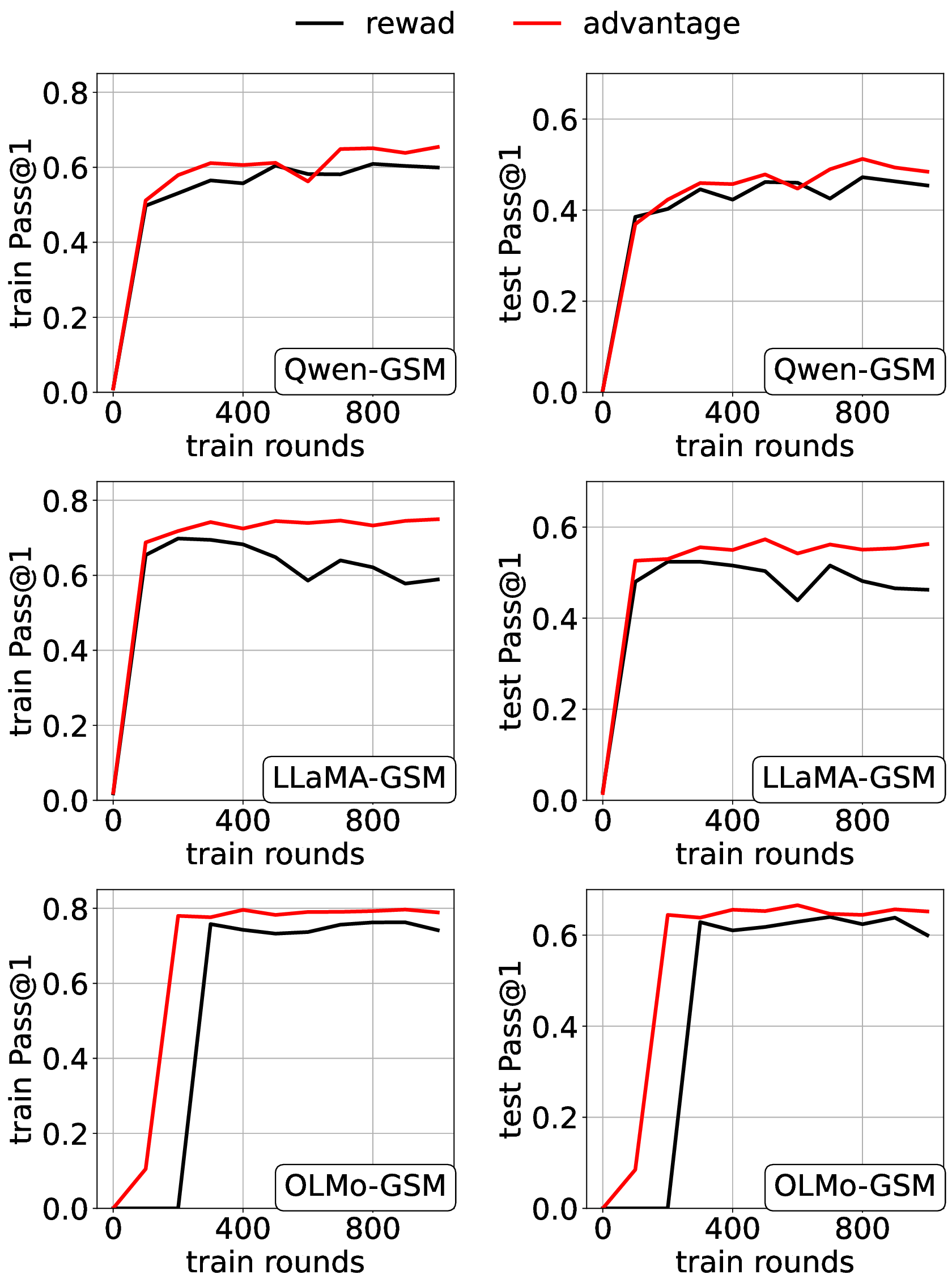}
}
\caption{Impact of advantage function on GSM8K dataset.}
\label{fig:adv-GSM}
\end{minipage}
\hfill
\begin{minipage}[t]{0.49\linewidth}
\centering
\subfigure{
  \includegraphics[width=0.95\linewidth]{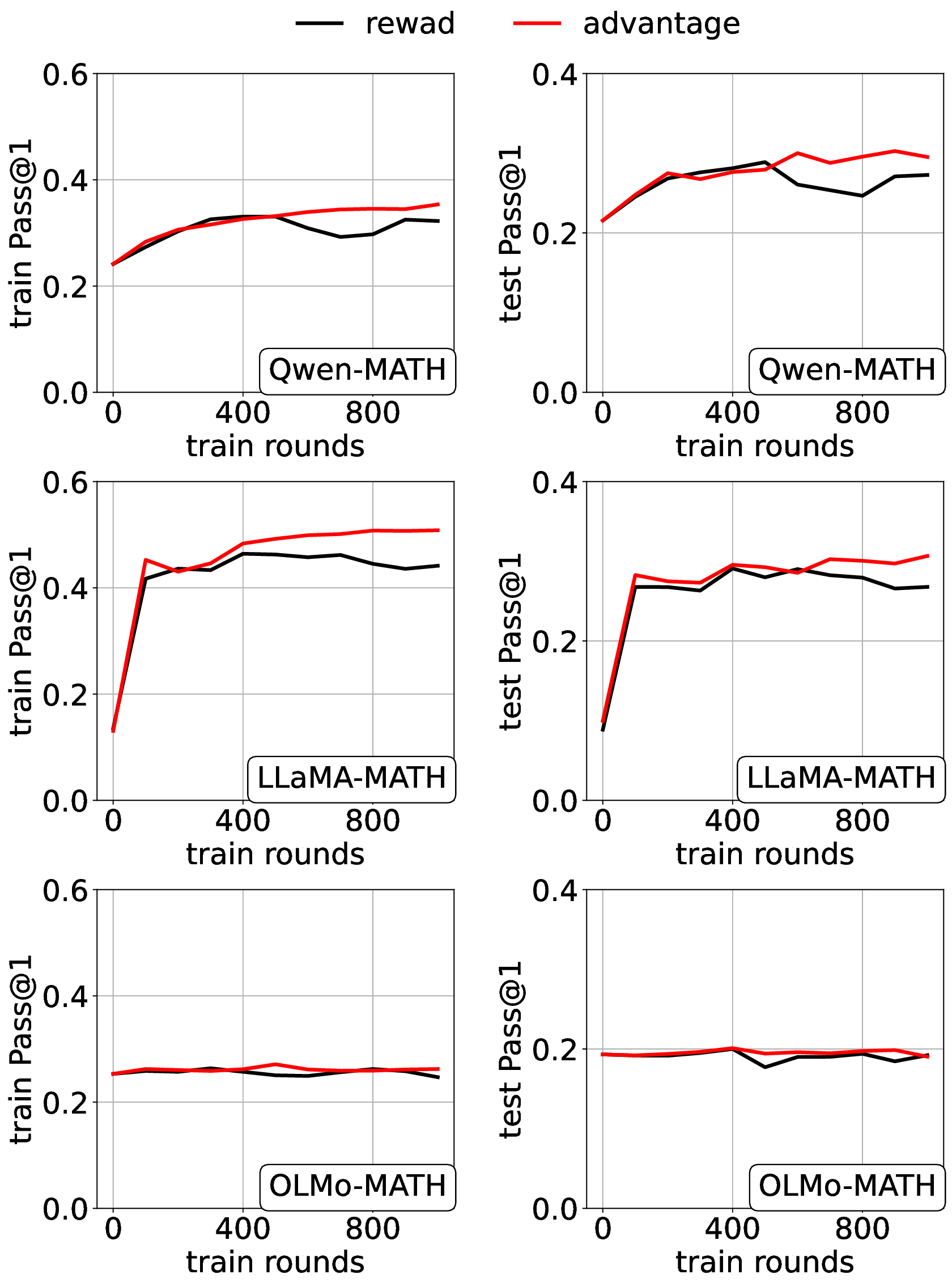}
}
\caption{Impact of advantage function on MATH dataset.}
\label{fig:adv-MATH}
\end{minipage}
\end{figure}

Figure \ref{fig:adv-GSM}
shows the improvement in Pass@1 by enabling advantage function on GSM8K dataset.
From the learning perspective,
one can observe that,
after enabling the advantage function of GRPO,
three models show small improvements in train Pass@1.
More specifically, 
the train Pass@1 of Qwen, LLaMA, and OLMo
improves by $0.05$ (from $0.60$ to $0.65$), $0.16$ (from $0.59$ to $0.75$), and $0.05$ (from $0.74$ to $0.79$).
From the generalization perspective,
one can observe that,
after enabling the advantage function of GRPO,
three models show small improvements in test Pass@1.
More specifically, 
the test Pass@1 of Qwen, LLaMA, and OLMo
improves by $0.03$ (from $0.45$ to $0.48$), $0.10$ (from $0.46$ to $0.56$), and $0.05$ (from $0.60$ to $0.65$).

Figure \ref{fig:adv-MATH}
shows the improvement in Pass@1 by enabling advantage function on MATH dataset.
From the learning perspective,
one can observe that,
after enabling the advantage function of GRPO,
Qwen and LLaMA show small improvements in train Pass@1, while OLMo shows almost no improvement during training.
More specifically, 
the train Pass@1 of Qwen, LLaMA, and OLMo
improves by $0.03$ (from $0.32$ to $0.35$), $0.07$ (from $0.44$ to $0.51$), and $0.01$ (from $0.25$ to $0.26$).
From the generalization perspective,
one can observe that,
after enabling the advantage function of GRPO,
Qwen and LLaMA also show small improvements in test Pass@1, while OLMo also shows almost no improvement.
More specifically, 
the test Pass@1 of Qwen, LLaMA, and OLMo
improves by $0.03$ (from $0.27$ to $0.30$), $0.04$ (from $0.27$ to $0.31$), 
and $0.00$ (from $0.19$ to $0.19$).

{\bf Key insights.}  
By enabling the advantage function:
1) The marginal improvement of test Pass@1 
varies from $0.00$ to $0.10$ across six model-dataset pairs,
where OLMo-MATH is $0.00$ (from $0.19$ to $0.19$),
and LLaMA-GSM is $0.10$ (from $0.46$ to $0.56$).
2) When the minimalist baseline in Section \ref{sec:pipeline} fails to improve the test Pass@1, 
enabling the advantage function also does not, such as OLMo-MATH.

\subsection{Exp3: the Role of Scaling Rollouts}
\label{sec:rollout}
We extend the setting of Section \ref{sec:adv} to study the role of scaling rollouts.  
The number of rollouts is varied among $1,8,16,32,64$, 
where the advantage function of GRPO is enabled.
Note that when the number of rollouts equals one,
to ensure a fair comparison,
we enable the advantage function following the same method in Section \ref{sec:adv}.

\begin{figure}[!htb]
\centering
\begin{minipage}[t]{0.49\linewidth}
\centering
\subfigure{
  \includegraphics[width=0.95\linewidth]{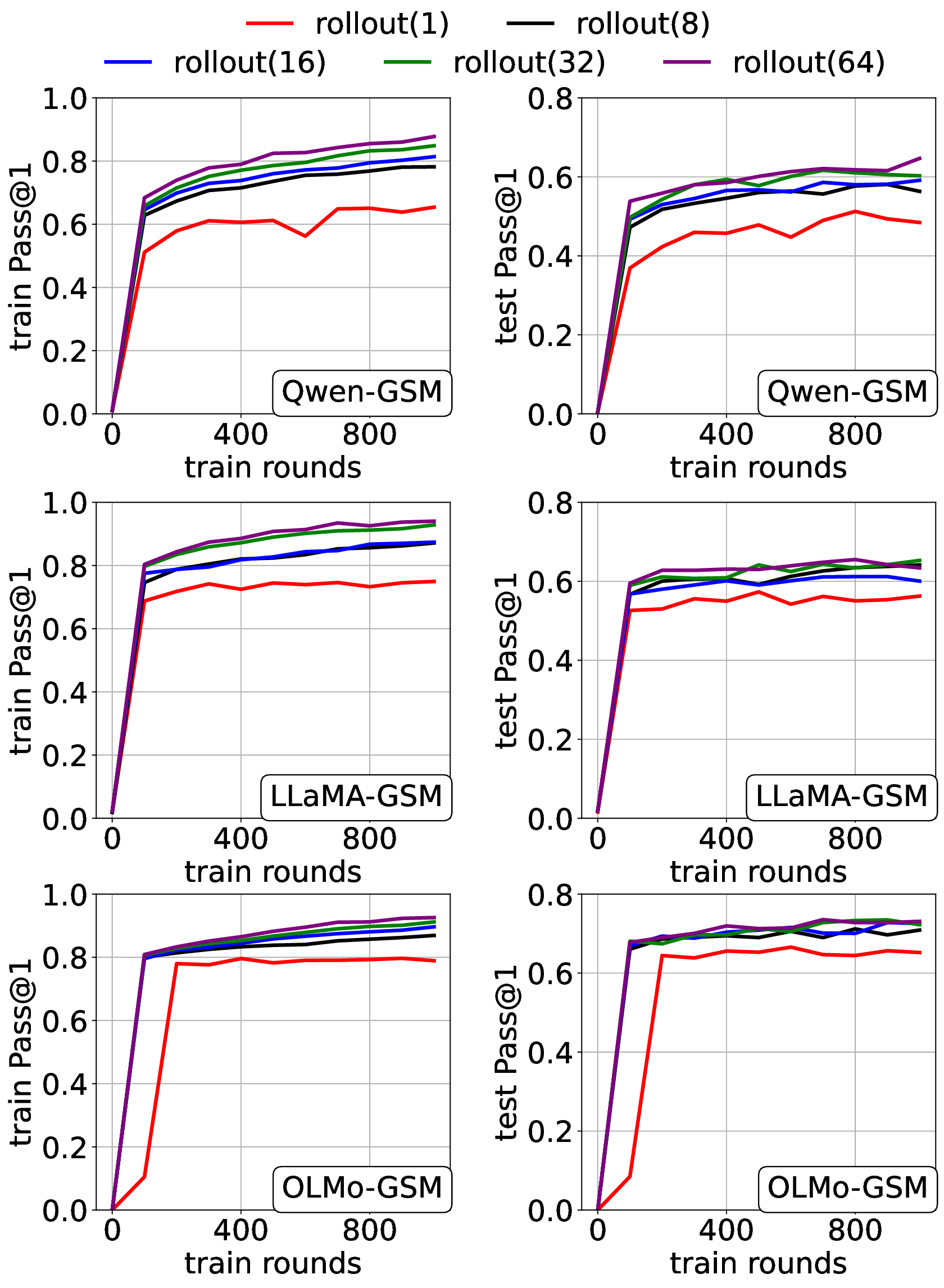}
}
\caption{Impact of scaling rollouts on GSM8K dataset.}
\label{fig:rollout-GSM}
\end{minipage}
\hfill
\begin{minipage}[t]{0.49\linewidth}
\centering
\subfigure{
  \includegraphics[width=0.95\linewidth]{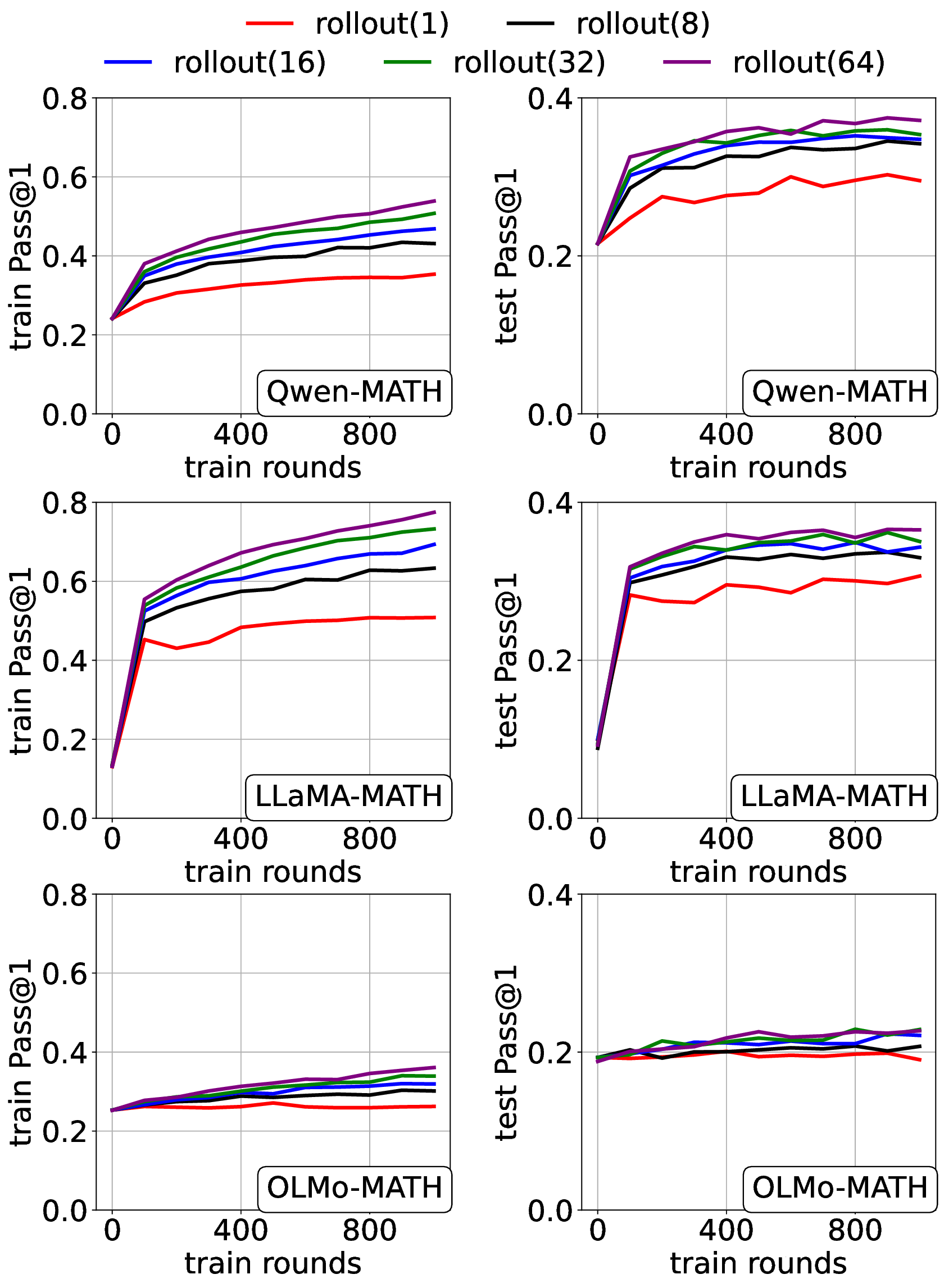}
}
\caption{Impact of scaling rollouts on MATH dataset.}
\label{fig:rollout-MATH}
\end{minipage}
\end{figure}

Figure \ref{fig:rollout-GSM}
shows the improvement in Pass@1 by scaling the number of rollouts on GSM8K dataset.
From the learning perspective,
one can observe that,
by increasing the number of rollouts from $1$ to $64$,
each base model shows small improvements in train Pass@1.
More specifically, 
the train Pass@1 of Qwen, LLaMA, and OLMo
improves by $0.23$ (from $0.65$ to $0.88$), $0.19$ (from $0.75$ to $0.94$), 
and $0.14$ (from $0.79$ to $0.93$).
From the generalization perspective,
one can observe that,
by increasing the number of rollouts from $1$ to $64$,
each base model shows small improvements in test Pass@1.
More specifically, 
the test Pass@1 of Qwen, LLaMA, and OLMo
improves by $0.17$ (from $0.48$ to $0.65$), $0.07$ (from $0.56$ to $0.63$), and $0.08$ (from $0.65$ to $0.73$).

Figure \ref{fig:rollout-MATH}
shows the improvement in Pass@1 by scaling the number of rollouts on MATH dataset.
From the learning perspective,
one can observe that,
by increasing the number of rollouts from $1$ to $64$,
each base model shows small improvements in train Pass@1.
More specifically, 
the train Pass@1 of Qwen, LLaMA, and OLMo
improves by $0.19$ (from $0.35$ to $0.54$), $0.26$ (from $0.51$ to $0.77$), and $0.10$ (from $0.26$ to $0.36$).
From the generalization perspective,
one can observe that,
by increasing the number of rollouts from $1$ to $64$,
each base model shows small improvements in test Pass@1.
More specifically, 
the test Pass@1 of Qwen, LLaMA, and OLMo
improves by $0.07$ (from $0.30$ to $0.37$), $0.06$ (from $0.31$ to $0.37$), and $0.04$ (from $0.19$ to $0.23$).

{\bf Key insights.}
By scaling the number of rollouts from $1$ to $64$:
1) The marginal improvement of test Pass@1 
varies from $0.04$ to $0.17$ across six model-dataset pairs,
where OLMo-MATH is $0.04$ (from $0.19$ to $0.23$),
and Qwen-GSM8K is $0.17$ (from $0.48$ to $0.65$);
2) When the minimalist baseline in Section \ref{sec:pipeline} does not improve the test Pass@1,
scaling rollouts only improves the test Pass@1 slightly by $0.04$ (from $0.19$ to $0.23$), such as OLMo-MATH.

\subsection{Exp4: the Role of Scaling Batch Size}
\label{sec:batch}
To investigate the role of batch size, 
we extend the setting of Section \ref{sec:rollout} 
by varying the batch size from $32$ to $128$,
while fixing the number of rollouts at eight.

\begin{figure}[!htb]
\centering
\begin{minipage}[t]{0.49\linewidth}
\centering
\subfigure{
  \includegraphics[width=0.95\linewidth]{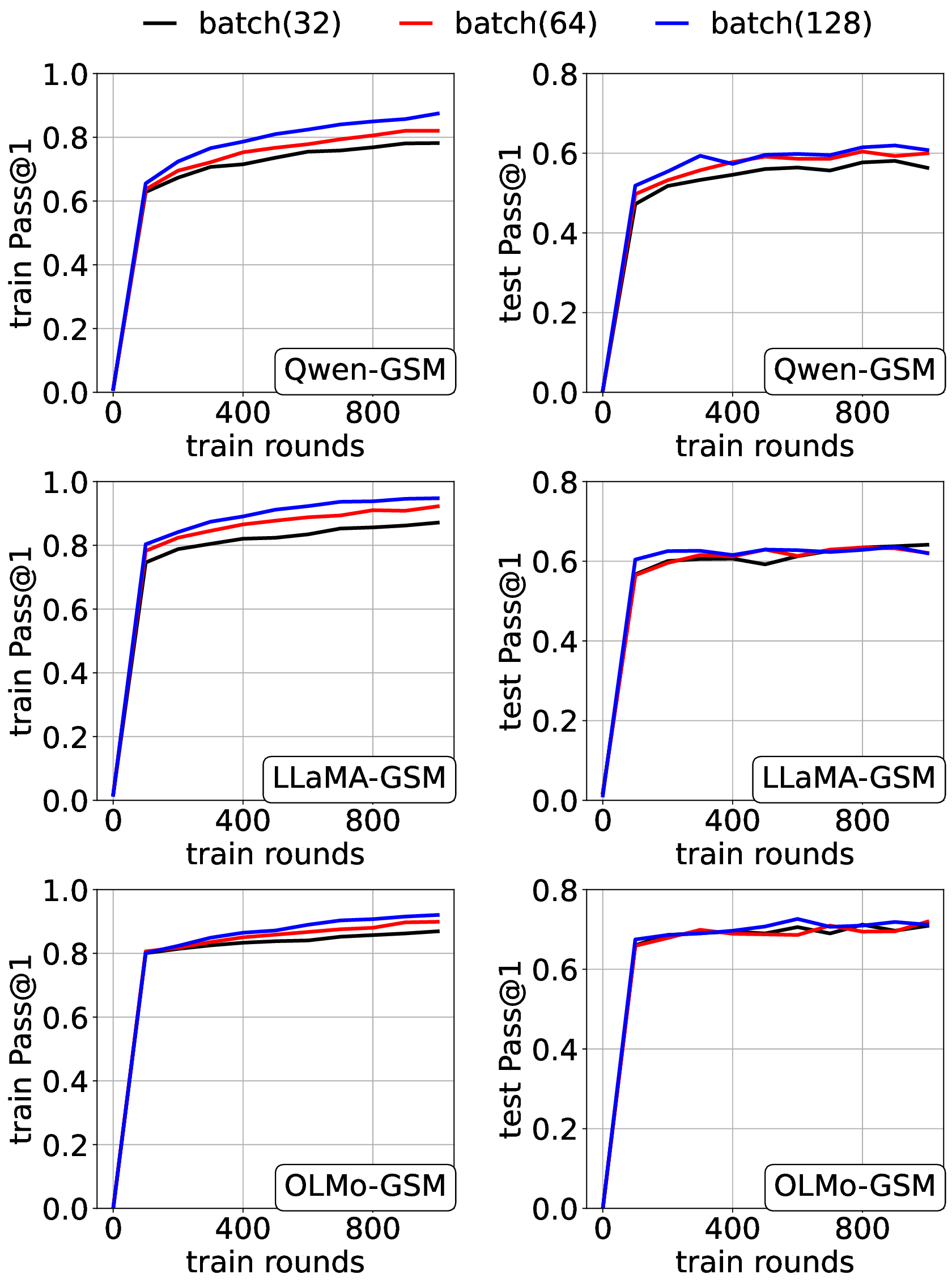}
}
\caption{Impact of scaling batch size on GSM8K dataset.}
\label{fig:batch-GSM}
\end{minipage}
\hfill
\begin{minipage}[t]{0.49\linewidth}
\centering
\subfigure{
  \includegraphics[width=0.95\linewidth]{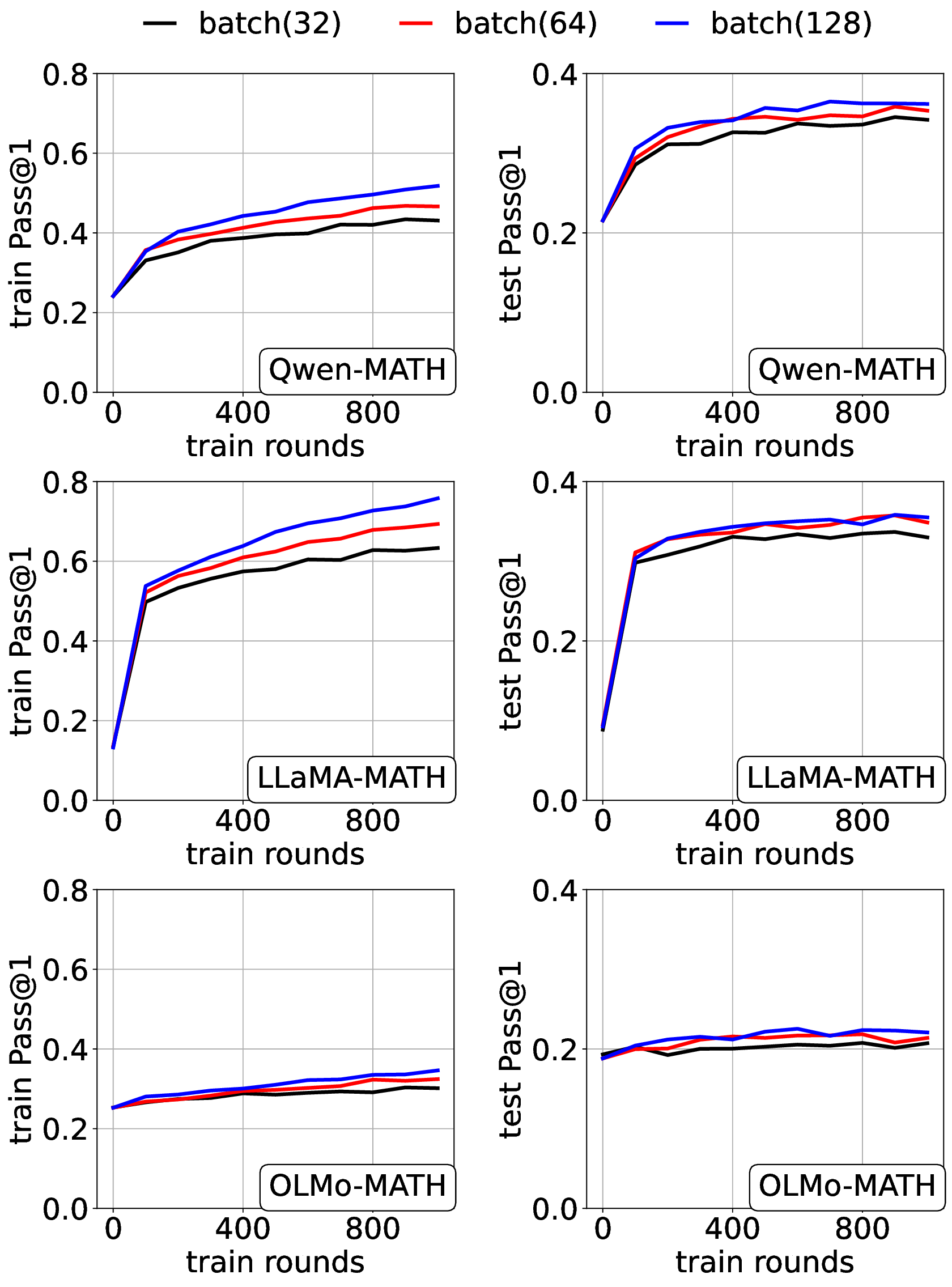}
}
\caption{Impact of scaling batch size on MATH dataset.}
\label{fig:batch-MATH}
\end{minipage}
\end{figure}

Figure \ref{fig:batch-GSM}
shows the improvement in Pass@1 by scaling the batch size on GSM8K dataset.
From the learning perspective,
one can observe that,
by increasing the batch size from $32$ to $128$,
each base model shows small improvements in train Pass@1.
More specifically, 
the train Pass@1 of Qwen, LLaMA, and OLMo
improves by 
$0.09$ (from $0.78$ to $0.87$), $0.08$ (from $0.87$ to $0.95$), and $0.05$ (from $0.87$ to $0.92$).
From the generalization perspective,
one can observe that,
by increasing the batch size from $32$ to $128$,
each base model shows small improvements in test Pass@1.
More specifically, 
the test Pass@1 of Qwen, LLaMA, and OLMo
improves by 
$0.05$ (from $0.56$ to $0.61$), $-0.02$ (from $0.64$ to $0.62$), and $0.00$ (from $0.71$ to $0.71$).

Figure \ref{fig:batch-MATH}
shows the improvement in Pass@1 by scaling the batch size on MATH dataset.
From the learning perspective,
one can observe that,
by increasing the batch size from $32$ to $128$,
each base model shows small improvements in train Pass@1.
More specifically, 
the train Pass@1 of Qwen, LLaMA, and OLMo
improves by $0.09$ (from $0.43$ to $0.52$), $0.13$ (from $0.63$ to $0.76$), and $0.05$ (from $0.30$ to $0.35$).
From the generalization perspective,
one can observe that,
by increasing the batch size from $32$ to $128$,
each base model shows small improvements in test Pass@1.
More specifically, 
the test Pass@1 of Qwen, LLaMA, and OLMo
improves by $0.02$ (from $0.34$ to $0.36$), $0.03$ (from $0.33$ to $0.36$), and $0.01$ (from $0.21$ to $0.22$).

{\bf Key insights.}
By scaling the batch size from $32$ to $128$:
1) The marginal improvement of test Pass@1 
varies from $0.00$ to $0.05$ across six model-dataset pairs,
where OLMo-GSM is $0.00$(from $0.71$ to $0.71$),
and Qwen-GSM is $0.05$(from $0.56$ to $0.61$).
2) When the minimalist baseline in Section \ref{sec:pipeline} does not improve the test Pass@1,
scaling batch size only improves the test Pass@1 slightly by $0.01$ (from $0.21$ to $0.22$), 
such as OLMo-MATH.

\subsection{EXP5: Batch Size vs. Rollouts Tradeoffs}
\label{sec:tradeoff}
We extend the setting of Section \ref{sec:batch} to study the 
batch size vs. rollouts tradeoffs.  
We fix (batch size) $\times$ (number of rollouts) $=256$,
and vary batch size over ${= 1, 2, 4, 8, 16, 32, 64, 128, 256}$
with corresponding numbers of rollout  
${=256,128,64,32,16,8,4,2,1}$.

\begin{figure}[!htb]
\centering
\begin{minipage}[t]{0.49\linewidth}
\centering
\subfigure{
  \includegraphics[width=0.95\linewidth]{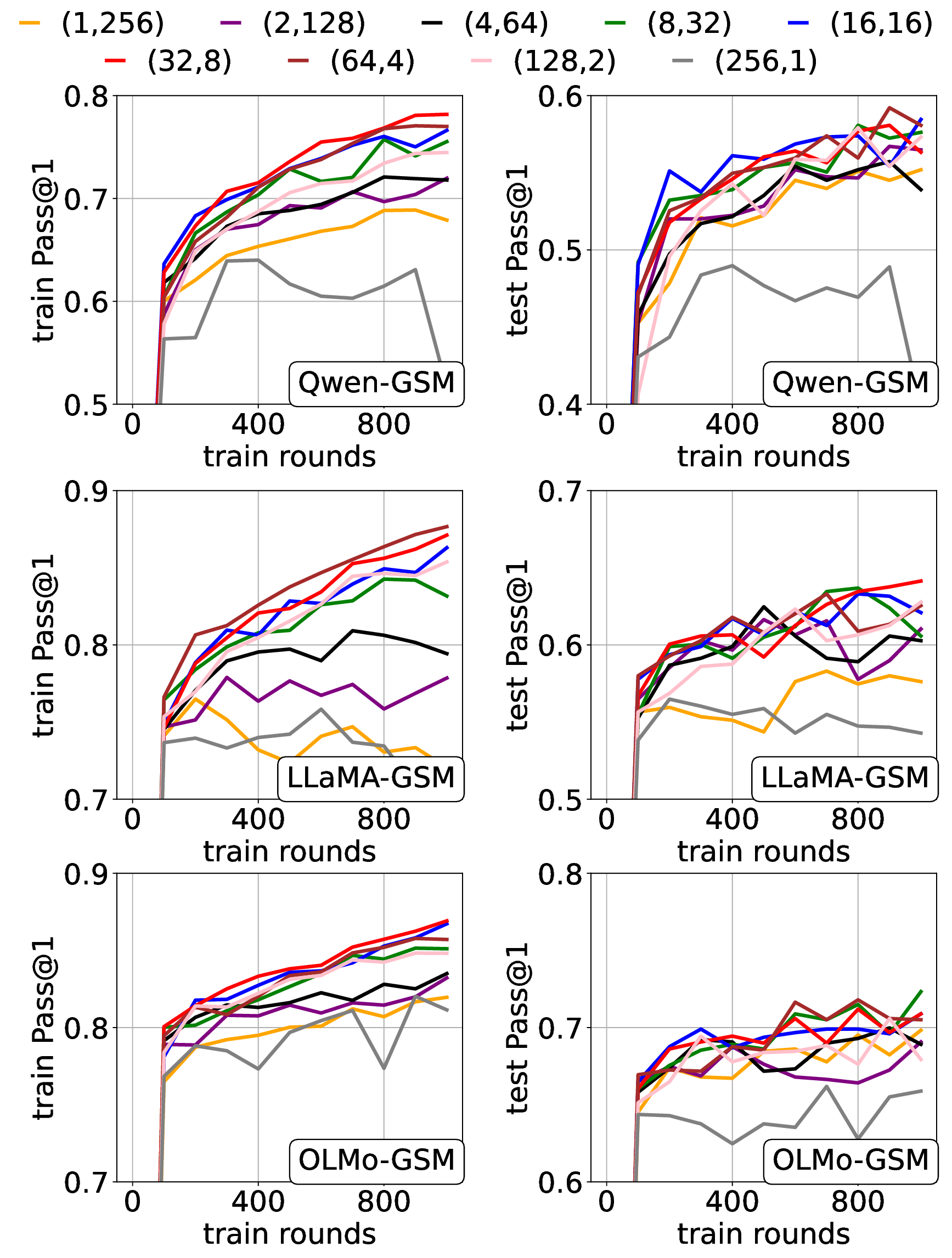}
}
\caption{Batch size vs. rollouts tradeoffs on GSM8K dataset.}
\label{fig:tradeoff-GSM}
\end{minipage}
\hfill
\begin{minipage}[t]{0.49\linewidth}
\centering
\subfigure{
  \includegraphics[width=0.95\linewidth]{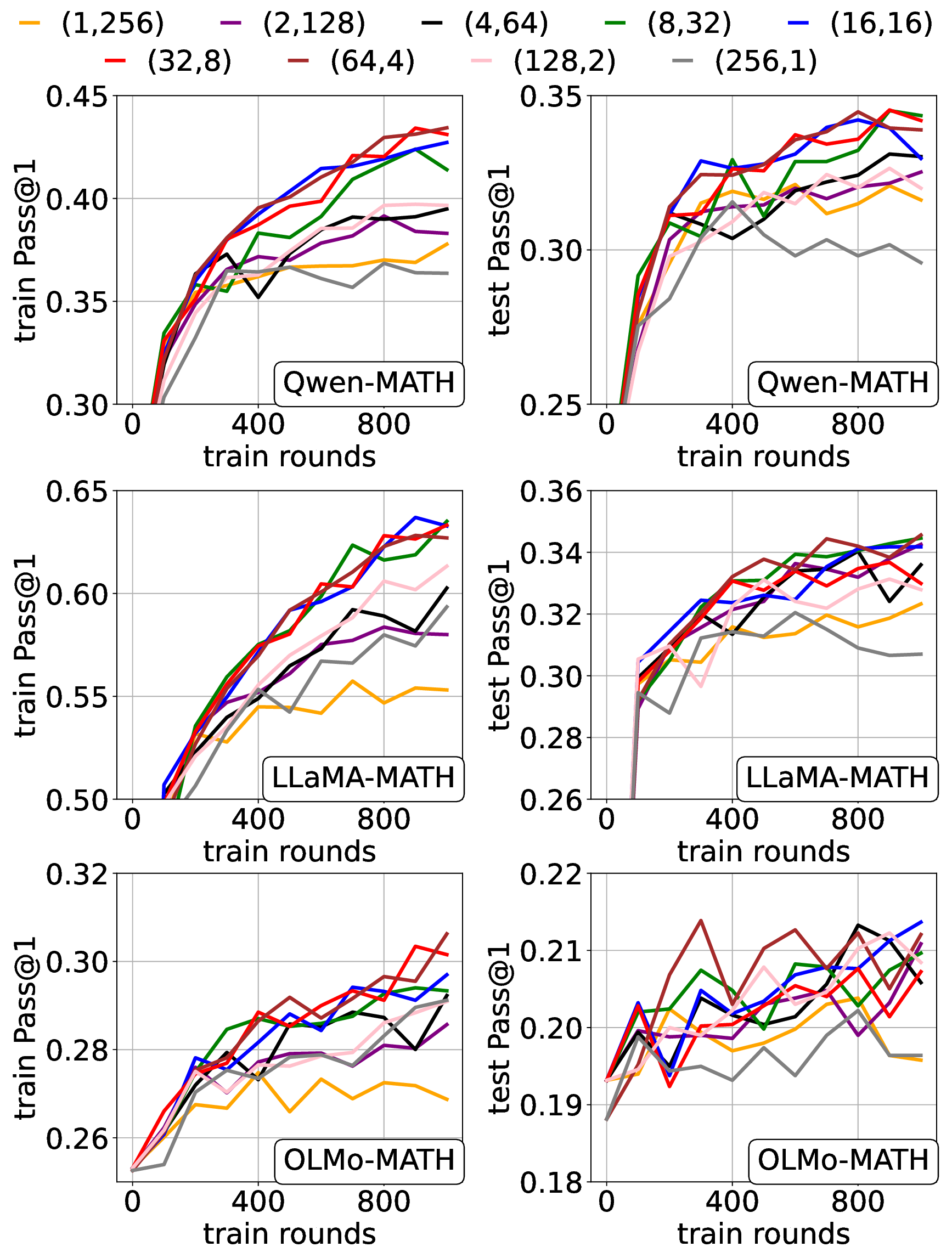}
}
\caption{Batch size vs. rollouts tradeoffs on MATH dataset.}
\label{fig:tradeoff-MATH}
\end{minipage}
\end{figure}

Figure \ref{fig:tradeoff-GSM} shows batch size vs. rollouts tradeoffs on GSM8K dataset.
From the learning perspective,
one can observe that,
different matches exhibit clear performance gains in train Pass@1.
More specifically, 
compared with $(256,1)$,
the train Pass@1 of Qwen, LLaMA, and OLMo with $(32,8)$ 
improves by $0.26$ (from $0.52$ to $0.78$), $0.16$ (from $0.71$ to $0.87$), and $0.06$ (from $0.81$ to $0.87$).
From the generalization perspective,
one can observe that,
different matches exhibit clear performance gains in test Pass@1.
More specifically, 
compared with $(256,1)$,
the test Pass@1 of Qwen, LLaMA, and OLMo with $(32,8)$ 
improves by $0.16$ (from $0.40$ to $0.56$), $0.10$ (from $0.54$ to $0.64$), and $0.05$ (from $0.66$ to $0.71$).

Figure \ref{fig:tradeoff-MATH} shows batch size vs. rollouts tradeoffs on MATH dataset.
From the learning perspective,
one can observe that,
different matches exhibit clear performance gains in train Pass@1.
More specifically, 
compared with $(256,1)$,
the train Pass@1 of Qwen, LLaMA, and OLMo with $(32,8)$ 
improves by $0.07$ (from $0.36$ to $0.43$), $0.04$ (from $0.59$ to $0.63$), and $0.01$ (from $0.29$ to $0.30$).
From the generalization perspective,
one can observe that,
different matches exhibit clear performance gains in test Pass@1.
More specifically, 
compared with $(256,1)$,
the test Pass@1 of Qwen, LLaMA, and OLMo with $(32,8)$ 
improves by $0.04$ (from $0.30$ to $0.34$), $0.02$ (from $0.31$ to $0.33$), and $0.01$ (from $0.20$ to $0.21$).

{\bf Key insights.}
By fixing (batch size) $\times$ (number of rollouts) $=256$, 
1) There is a tradeoff between batch size and number of rollouts,
i.e.,
compared with $(256,1)$, the improvement of test Pass@1 with $(32,8)$ 
varies from $0.01$ to $0.16$ across six model-dataset pairs,
where OLMo-MATH is $0.01$ (from $0.20$ to $0.21$),
and Qwen-GSM is $0.16$ (from $0.40$ to $0.56$).
2) When the minimalist baseline in Section \ref{sec:pipeline} does not improve the test Pass@1,
the tradeoff problem is small, with an improvement of only $0.01$ (from $0.20$ to $0.21$), such as OLMo-MATH.

\subsection{Exp6: Attaining Optimal Tradeoff Via Replay}
\label{sec:replay}

Section \ref{sec:tradeoff} illustrates the tradeoff problem between the number of rollouts and batch size
in GRPO learning and generalization.  
In practice, searching for the optimal tradeoff is computationally expensive.  
We design a replay strategy to attain the optimal tradeoff approximately.
For each question,
we maintain a replay buffer that stores the most recent rollouts,
and compute the advantage function following GRPO \cite{shao2024deepseekmath}.
Note that to ensure a fair comparison,
only the advantage corresponding to the current rollouts is used to compute the policy gradient, 
while replayed rollouts are used solely to support the advantage estimation of the current rollouts.

Section \ref{sec:tradeoff} shows that 
under the fixed (batch size) $\times$ (number of rollouts) $=256$,
the train Pass@1 and test Pass@1 of match $(32,8)$ are the optimal or close to the optimal 
across different model-dataset pairs.
Thus, we set the match $(32,8)$ as the optimal baseline.
We extend the setting of Section \ref{sec:tradeoff} 
to evaluate the effect of our replay strategy.
We fix (batch size) $\times$ (current rollouts $+$ replay rollouts) $=256$,
set batch size as thirty-two and
(current rollouts $+$ replay rollouts) $=8$,
and vary replay rollouts over ${= 7,6,4}$
with corresponding current rollouts  
${=1,2,4}$.

\begin{figure}[!htb]
\centering
\begin{minipage}[t]{0.49\linewidth}
\centering
\subfigure{
  \includegraphics[width=0.95\linewidth]{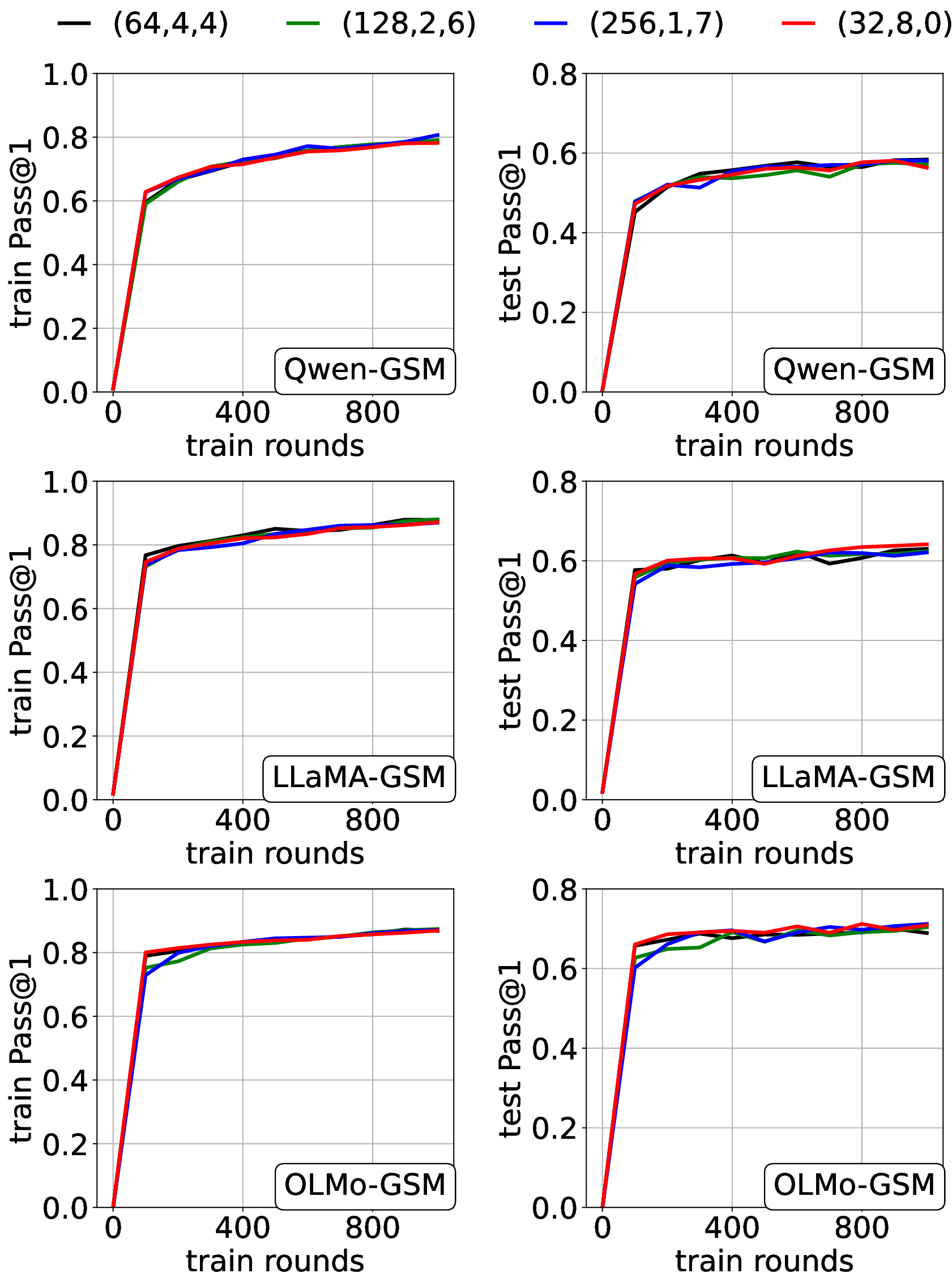}
}
\caption{A replay strategy to attain the optimal batch size vs. rollouts tradeoff on GSM8K dataset.}
\label{fig:replay-GSM-optimal}
\end{minipage}
\hfill
\begin{minipage}[t]{0.49\linewidth}
\centering
\subfigure{
  \includegraphics[width=0.95\linewidth]{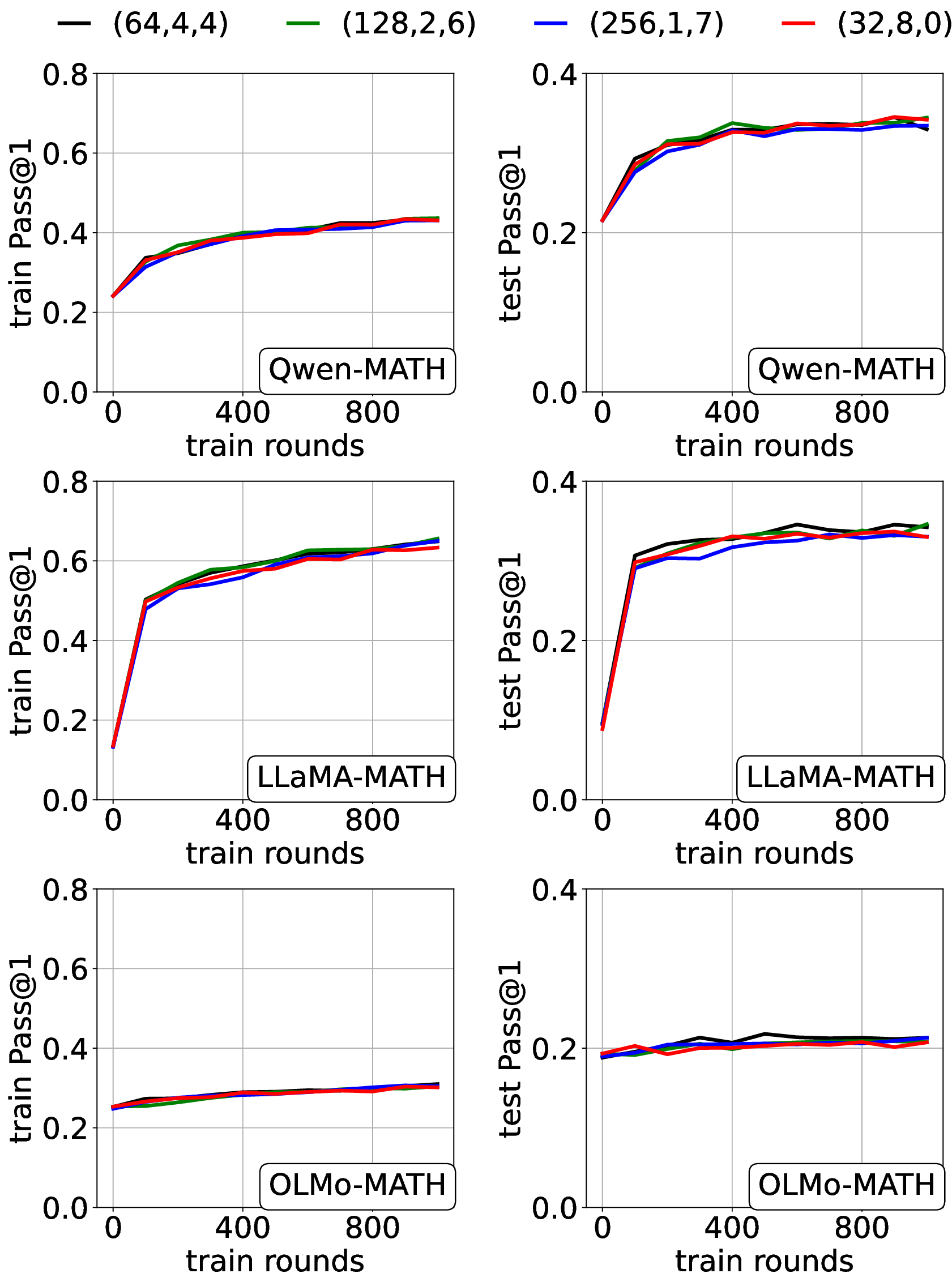}
}
\caption{A replay strategy to attain the optimal batch size vs. rollouts tradeoff on MATH dataset.}
\label{fig:replay-MATH-optimal}
\end{minipage}
\end{figure}

Figure \ref{fig:replay-GSM-optimal}
shows our replay strategy of different tuples 
(batch size, current rollouts, replay rollouts) on GSM8K dataset.
From the learning perspective,
one can observe that 
the train Pass@1 curve of different replay tuples overlap with the optimal baselines $(32,8,0)$ across three models (Qwen, LLaMA, and OLMo).
This implies that
our replay strategy can attain learning performance comparable to the optimal baselines on GSM8K dataset.
From the generalization perspective,
one can observe that 
the test Pass@1 curve of different replay tuples overlap with the optimal baselines $(32,8,0)$ across three models (Qwen, LLaMA, and OLMo).
This implies that 
our replay strategy can attain generalization performance comparable to the optimal baselines on GSM8K dataset.

Figure \ref{fig:replay-MATH-optimal}
shows our replay strategy of different tuples 
(batch size, current rollouts, replay rollouts) on MATH dataset.
From the learning perspective,
one can observe that 
the train Pass@1 curve of different replay tuples overlap with the optimal baselines $(32,8,0)$ across three models (Qwen, LLaMA, and OLMo).
This implies that
our replay strategy can attain learning performance comparable to the optimal baselines on MATH dataset.
From the generalization perspective,
one can observe that 
the test Pass@1 curve of different replay tuples overlap with the optimal baselines $(32,8,0)$ across three models (Qwen, LLaMA, and OLMo).
This implies that 
our replay strategy can attain generalization performance comparable to the optimal baselines on MATH dataset.

{\bf Key insights.} 
1) Our replay strategy can attain test Pass@1 comparable to the optimal baselines ($32,8,0$)
across six model-dataset pairs.
2) When the minimalist baseline in Section \ref{sec:pipeline} does not improve the test Pass@1,
our replay strategy also does not, such as OLMo-MATH.

\subsection{Exp7: Performance Ceiling of Scaling}
\label{sec:scaling-replay}
We extend the setting of Section \ref{sec:replay}, 
and apply the replay strategy to study the performance ceiling of scaling batch size.  
To achieve this, we vary batch size among $256,512,1024,2048$.

\begin{figure}[!htb]
\centering
\begin{minipage}[t]{0.49\linewidth}
\centering
\subfigure{
  \includegraphics[width=0.95\linewidth]{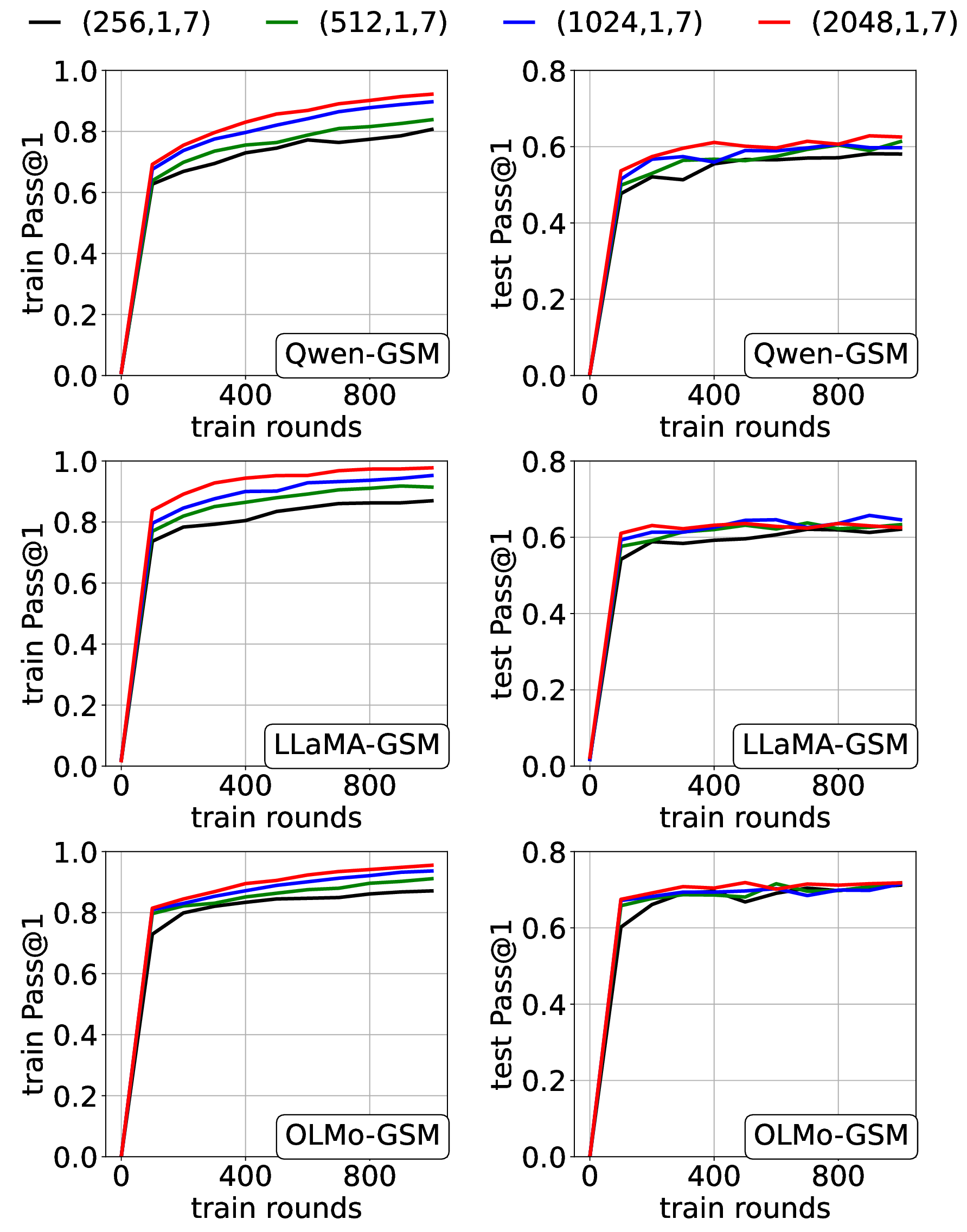}
}
\caption{Exploring the performance ceiling by scaling on GSM8K dataset.}
\label{fig:replay-batch-GSM}
\end{minipage}
\hfill
\begin{minipage}[t]{0.49\linewidth}
\centering
\subfigure{
  \includegraphics[width=0.95\linewidth]{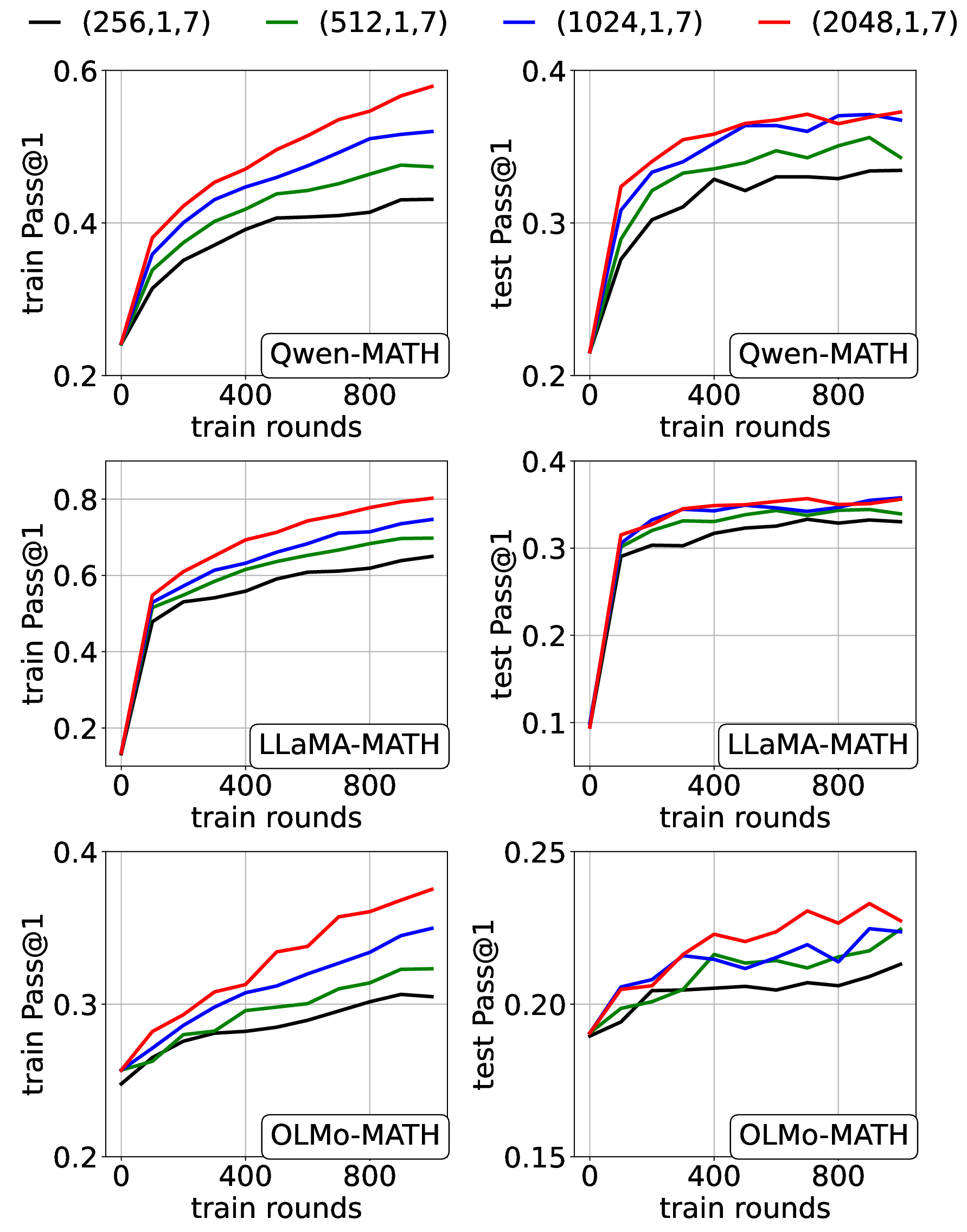}
}
\caption{Exploring the performance ceiling by scaling on MATH dataset.}
\label{fig:replay-batch-MATH}
\end{minipage}
\end{figure}

Figure \ref{fig:replay-batch-GSM}
shows our replay strategy by scaling the batch size on GSM8K dataset.
From the learning perspective,
one can observe that 
by increasing the batch size from $256$ to $2048$, 
each base model shows small improvements in train Pass@1.
More specifically, 
the train Pass@1 of Qwen, LLaMA, and OLMo
improves by 
$0.11$ (from $0.81$ to $0.92$), $0.11$ (from $0.87$ to $0.98$), and $0.09$ (from $0.87$ to $0.96$).
From the generalization perspective,
one can observe that 
by increasing the batch size from $256$ to $2048$, 
each base model shows small improvements in test Pass@1.
More specifically, 
the test Pass@1 of Qwen, LLaMA, and OLMo
improves by 
$0.05$ (from $0.58$ to $0.63$), $0.01$ (from $0.62$ to $0.63$), and $0.01$ (from $0.71$ to $0.72$).

Figure \ref{fig:replay-batch-MATH}
shows our replay strategy by scaling the batch size on MATH dataset.
From the learning perspective,
one can observe that 
by increasing the batch size from $256$ to $2048$, 
each base model shows small improvements in train Pass@1.
More specifically, 
the train Pass@1 of Qwen, LLaMA, and OLMo
improves by $0.15$ (from $0.43$ to $0.58$), $0.15$ (from $0.65$ to $0.80$), and $0.08$ (from $0.30$ to $0.38$).
From the generalization perspective,
one can observe that 
by increasing the batch size from $256$ to $2048$, 
each base model shows small improvements in test Pass@1.
More specifically, 
the test Pass@1 of Qwen, LLaMA, and OLMo
improves by $0.04$ (from $0.33$ to $0.37$), $0.03$ (from $0.33$ to $0.36$), and $0.02$ (from $0.21$ to $0.23$).

{\bf Key insights.}
By scaling the batch size of our replay strategy:
1) The marginal improvement of test Pass@1 
varies from $0.01$ to $0.05$ across six model-dataset pairs,
where OLMo-GSM is $0.01$ (from $0.71$ to $0.72$),
and Qwen-GSM is $0.05$ (from $0.58$ to $0.63$).
2) When the minimalist baseline in Section \ref{sec:pipeline} does not improve the test Pass@1,
scaling batch size of our replay strategy only improves the test Pass@1 slightly by $0.02$ (from $0.21$ to $0.23$), such as OLMo-MATH.

\section{Conclusion}

This paper studied the design choices of reinforcement fine-tuning.  
A minimalist baseline 
for disentangling factors is constructed: 
one rollout per query in each round, 
the outcome reward serving as the learning signal 
without any advantage trick, and a batch size of thirty-two.   
This baseline is closely related to 
batched contextual bandit learning, 
which facilitates experimental analysis.  
Centering around this baseline, 
an experimental pipeline is designed, 
examining the marginal gains of factors like advantage, number of rollouts, etc.  
Experiments on three base models and two datasets,  
not only reveal new insights into the role of various design choices 
on learning and generalization dynamics, 
but also identify critical ones that deserve more effort.

\section*{Impact Statement}
This paper presents work whose goal is to advance the field of reinforcement fine-tuning. 
There are many potential societal consequences of our work, 
none of which we feel must be specifically highlighted here.

\bibliographystyle{named}
\bibliography{Reference}

\end{document}